\documentclass[letterpaper, 10 pt, conference]{ieeeconf}  % Comment this line out if you need a4paper

\IEEEoverridecommandlockouts                              % This command is only needed if 
\usepackage{amsmath} % assumes amsmath package installed
\usepackage{amssymb}  % assumes amsmath package installed

\usepackage{algorithm}
\usepackage{algpseudocode}

\usepackage{svg}
\usepackage{multirow}

 \usepackage{graphicx}
 \usepackage{comment}

\usepackage{xcolor}

\usepackage{listings}
\usepackage{xcolor}

\lstdefinelanguage{json}{
    basicstyle=\ttfamily\small,
    numbers=none,
    breaklines=true,
    frame=single,
    showstringspaces=false
}

\usepackage{bm}
\title{\LARGE \bf
Fail-RAG : A Retrieval Augmented Generation Informed Framework for Robot Failure Identification}

\author{ Ameya Salvi$^{1}$ and Jie Hu$^{1}$ % <-this % stops a space
\thanks{$^{1}$ Authors are with Hitachi America, Ltd., Michigan, U.S.A.,
       {\tt\small {ameya.salvi, jie.hu}@hal.hitachi.com}}%
}
\begin{document}

\maketitle
\thispagestyle{empty}
\pagestyle{empty}

%%%
% Abstract + Introduction: 1 page
% Related work: 3/4 page
% Methods: 1
% Exp: 2.5 pages
% References: 3/4 page
%%%
%%%%%%%%%%%%%%%%%%%%%%%%%%%%%%%%%%%%%%%%%%%%%%%%%%%%%%%%%%%%%%%%%%%%%%%%%%%%%%%%

\begin{abstract}
Industry automation is witnessing an evolution in robotics driven by both technological breakthroughs and societal changes: progress towards generalist robots, embodied and physical artificial intelligence (AI), and increasing labor shortage in manufacturing.
%While highly repetitive tasks in controlled environments have largely been automated, functioning reliably and adaptively for mixed tasks in dynamic environments is still challenging for robots.
An intelligent autonomous robot needs to not only \textit{act} according to planned motions but also \textit{react} to any unexpected events. In this study, we focus on such unexpected events in warehouses where robots are used for material handling. Specifically, we refer to any unexpected events as \textit{failures} and develop methods to detect robot operations related failures. Rule-based detection methods may break since the form of failures could change due to the dynamic nature of both environments and tasks. We propose \textbf{Fail-RAG}, a Retrieval Augmented Generation (RAG)-based failure detection framework where failure images and context information are embedded and queried against a failure database by calculating their similarities. Vision-Language Models (VLMs) are further used to analyze failures and provide details by following our instruction template. We evaluated the performance of Fail-RAG by conducting both simulation and physical experiments using fixed robot arms and a mobile manipulator for multiple tasks that are common in warehouse automation. Fail-RAG achieved 25 percentage point higher failure detection accuracy on average across five types of robot operations compared to using off-the-shelf VLMs, indicating its effectiveness for real-world failure detection.
\end{abstract}

%%%%%%%%%%%%%%%%%%%%%%%%%%%%%%%%%%%%%%%%%%%%%%%%%%%%%%%%%%%%%%%%%%%%%%%%%%%%%%%%
\section{INTRODUCTION}

The focus of automation has been shifting to more complex tasks and those that are natural to human but difficult to machines, examples exist from handling large volume of packages with great variety in logistics and warehouses\cite{yano2022}, to outdoor deliveries\cite{kim2024}, assistance in hospitals\cite{novin2018}, and household tasks\cite{kazhoyan2021}. While several automation processes have been perfected over decades, the surge in inexpensive hardware and development of cutting edge software has led to a wider attraction and adoption of robot technologies in increasingly challenging and unstructured environments. Conventionally, automation operations necessitate implementation of failure and anomaly identification frameworks for enhancing the reliability and safety of the operation. While several methods implementing specialized hardware and rule-based softwares have contributed to this area, the adoption of these method for more general purpose robots is significantly challenging due to complexities in scaling hardware and rule-base software. Inasmuch, vision language models (VLMs) provide a viable solution for monitoring robot operations purely based on camera vision information, allowing for their flexible application in a large number of robot operations.

Large language models (LLMs) have gained tremendous popularity in recent years for the development of \textit{agentic} systems (systems and workflows that can be automated to operate with minimum human supervision. Within the LLM ecosystem, VLMs have gained attention within the robotics community to utilize video and image data by leveraging multi-modal architectures, thus finding applications in scene understanding, behavior planning and robot control. More crucially, VLMs generate a human comprehensible output in the form of natural language which thus forms a critical bridge in understating robot's reasoning capabilities. Furthermore, the adoption of camera vision (with or without natural language) for model inference now allows to scale the operations in a wide variety of robotics tasks without installation of task specific sensors or development of operation specific algorithms. To this end, VLMs provide a promising opportunity to be implemented as a failure and anomaly identification mechanism in a variety of robot operations.

\begin{figure}
    \centering
    \includegraphics[width=\columnwidth]{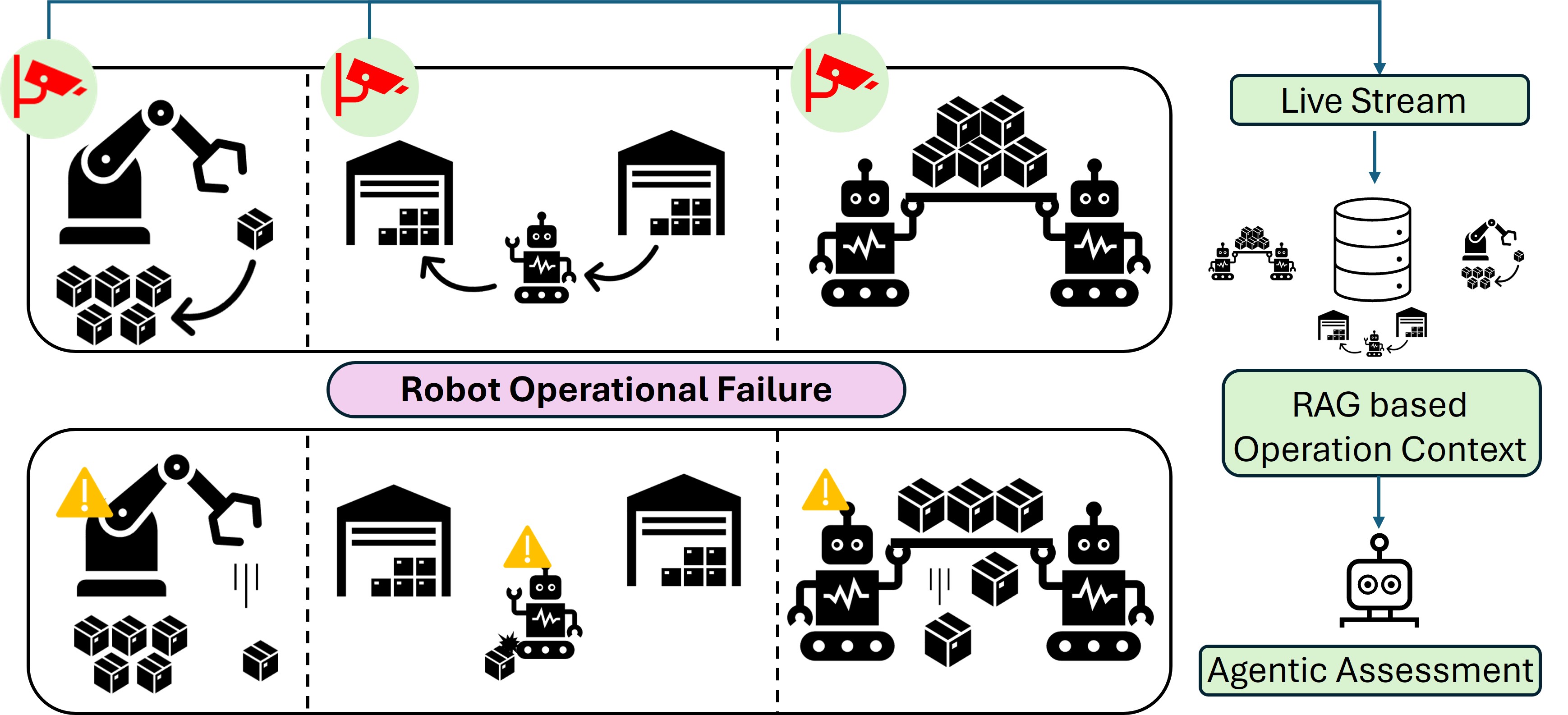}
    \caption{Overview of the proposed RAG-based failure detection framework. Multiple robot processes such as (a) object manipulation, (b) robot navigation, and (c) robot coordination are encoded as RAG context and used for real-time monitoring of robot operations using multi-modal language models.}
    \label{fig:overview}
\end{figure}

To this end, a RAG informed VLM framework, Fail-RAG, is proposed for identifying anomalies and failures in industrial robot operations. Figure~\ref{fig:overview} now captures the overall idea of the proposed framework. As compared to the existing state of the art, the frameworks does not require any fine-tuning for introducing new edge cases and can be implemented for real-time operations on significantly less compute. Furthermore, a statistical analysis for several parametric variations of the RAG formulation is presented to consolidate the framework's performance. More critically, deeper insights on the framework's performance driven by the nature of the vector embeddings has been presented in the analysis section.

Section~\ref{sec:related-work} reviews recent work in failure and anomaly detection in robotics domain. Section~\ref{sec:formulation} presents the problem formulation and the proposed methods. Section~\ref{sec:experiments} discusses the experiment design and results followed up by section~\ref{sec:analysis} providing insights in the operation of Fail-RAG. Finally, section~\ref{sec:discussion} provides conclusions and some future work directions.

\section{Related work}\label{sec:related-work}

Anomaly detection and Out-of-Distribution detection are two related research topics and methods developed in these field can be potentially applied for robot operation failures. While a rich set of literature indulges in adopting conventional or rule-based methods towards anomaly detection, they face challenges in scaling up to larger set of operations and their subsequent edge cases~\cite{VISINSKY1994FaultDetection,PETTERSSON2005Execution,Hwang2010FDIR}. To this end, we focus on recent study of using generative models for failure detection, such as LLMs, VLMs and Vision-Language-Action (VLAs).

%% Models that needing fine-tuning

%%%% Failure-identification only

A large body of work builds upon the conventional idea of finding discrepancies in camera vision information (images) as compared to a baseline or `normal' image. For example, failures where robot performs tasks that are different from instructions but semantically normal are considered as ``Semantic Misalignment Failures'' in~\cite{grislain_i-failsense_2025}. Detection of such failures is based on post-training VLMs using collected semantics misalignment dataset and adding classification heads. Similarly,~\cite{pacaud_guardian_2025} emphasizes the importance of failure data with fine-grained failure categories and reasoning, and a VLM training on multi-view images. As contrast to comparing images,~\cite{chong_robust_2026} proposes to construct a unified scene graph from multi-view scene images to provide full knowledge about the environment, which is further used to compare with an expected scene graph generated during planning to detect any failures. A similar framework of integrating ontologies, logical rules, and LLM-based planners is proposed for online failure identification and recovery for robotic operations in \cite{cornelio2024}. 

%%%% Failure-identification and recovery
While the body of work introduced above focuses on difference driven failure identification, a recent set of work goes a step further for failure identification and subsequent recovery. AHA ~\cite{duan_aha_2024}, proposes fine-tuning a language model on a failure dataset. Similarly, FailSafe~\cite{lin_failsafe_2025} is a finetuned VLM that detects failures and outputs executable actions to recover from detected failures. ViFailback was introduced in~\cite{zeng_diagnose_2025}, where visual symbols are generated to provide recovery guidance. SAFE, proposed in~\cite{gu_safe_2025} uses a single scalar to predict failures by learning the internal latent features.

Thus, a large set of contemporary works focus on fine-tuning existing model for failure identification or/and recovery. While powerful, the major drawback of these methods is (a) the necessity to capture vision-action failure data, and, (b) sourcing heavy compute resources for tuning vision-language models (which may sometimes go upwards of a few billion parameters).

%% Models that don't need fine-tuning
In contrast, several methods also explore the idea of identifying failures without the use of extensive datasets or fine-tuning. For example, anomalies are classified into robot-driven and environment-drive in~\cite{willibald_multimodal_2025}, and a mix-of-experts framework was proposed to detect anomalies from both sources by combining VLM and Gaussian-mixture models. VLMs are directly used as monitoring tooling during task execution in~\cite{ahmad_addressing_2024}. Further, skills for robot to address such failures are generated by VLMs and incorporated into the robot behavior tree. While interesting, the above works are limited by their capacity to identifying categorical failures (via outliers or deterministic behavior trees) and do not exploit the powerful capabilities of VLMs in providing semantic reasoning for the failures. Such a reasoning or insight can be critical in identifying effective means of intervention and failure corrections, especially in operations requiring human-robot collaborations.

Thus, the primary focus of our work tries to address two potential gaps in the existing literature:
\begin{itemize}
    \item Formulate a framework free of fine-tuning and thus can be adopted relatively easily within robot operation workflows.
    \item Utilizes the powerful language model capabilities in not only identifying the failures, but also provide insights on the potential cause of the failure.
\end{itemize}

%Distinct from all the above approaches for failure identification and recovery, it is also valuable to recognize efforts defining and categorizing what is meant by operational failures both in the past~\cite{}

%~\cite{wu_vulnerability_2025} offers a different perspective on LLM/VLM-controlled robots, revealing potential vulnerabilities due to sensitivities to input modalities. which might be an issue that VLM-based failure detection should consider as well.

\section{Formulation}\label{sec:formulation}

\begin{figure*}
    \centering
    \includegraphics[width=\textwidth]{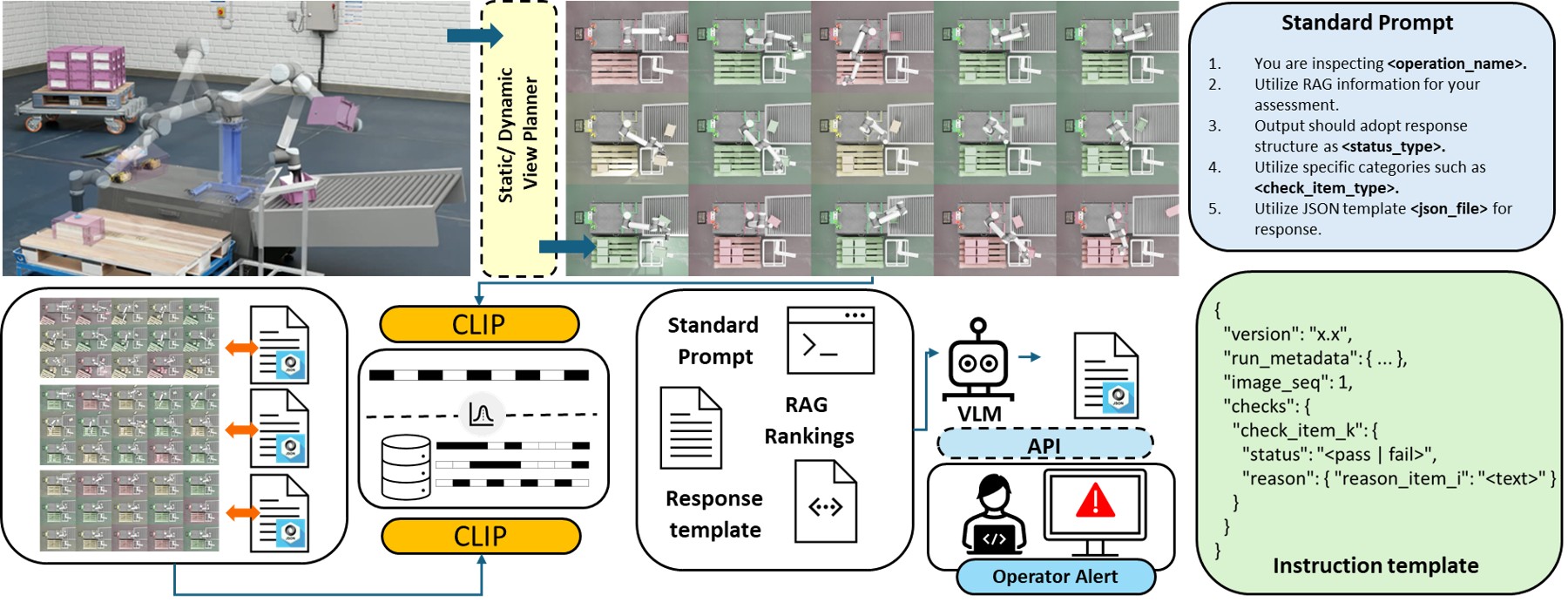}
    \caption{Overview of the proposed motion anomaly detection framework. A view planner captures frames from a video streaming object based on a one of the fixed frame rates ${5, 10, 15, 20, 25}~\text{Frames}$. The single image compiled of several frames is encoded as a $\mathbb{R}^{512}$ vector using the CLIP model. A pre-compiled RAG dataset tries to gather the relevant failure modes using one of the vector comparison metrics (cosine similarity, dot product, or, $\mathcal{L}_2$ distance). The ranked retrieval information is presented to a off-the-shelf VLM agent along with a standardized prompt and expected response template. The structures response can be integrated within existing API tools for human interpretation and subsequent operation intervention.}
    \label{fig:framework_pipeline}
\end{figure*}

Figure~\ref{fig:framework_pipeline} illustrates the flow of the proposed framework. Inspection camera video stream is received by an image processing layer which compiles the video frames as a single image containing time-sequence images. This images is provide to a vision language model along with (a) a process specific prompt and, (b) a process specific answer template. The provisioned answer template follows a structured format such as \verb|json| or \verb|yaml| that allows for seamless integration with an application programming interface (API).

\subsection{Retrieval Augmented Generation (RAG) Embeddings}
As discussed in section~\ref{sec:related-work} majority of the state-of-the art frameworks rely heavily on fine-tuning on failure scenarios requiring significant compute resources for VLM-finetuning. To avoid this, a minimalistic RAG based framework is proposed where failure scenarios are provided to the VLM as reference documents to predict operational failure. In this novel framework, the  pairs of time-sequenced images with their respective ground truth scenarios are embedded within a memory stack. For efficient processing of multi-modal data~(image and text), Contrastive Language-Image Pre-Training (CLIP) encoder is used~\cite{radford2021learningtransferablevisualmodels}. As illustrated in Fig.~\ref{fig:framework_pipeline}, the real-time image data is processed as CLIP embeddings and is used to retrieve the relevant information from the memory stash by utilizing comparison metrics between the vector embeddings. 

As witnessed in contemporary works, different metrics used within the RAG formulation to measure the distance between the indexed embeddings and the query embeddings lead to different performance~\cite{Elkiran2025EvaRAG}. Typical strategies for making the choice of the measurement metric depends on the data that is used to create the embeddings (such as text, graphics or audio). In this work, since we are dealing with multi-modal embeddings, we have chosen three metrics to compare and investigate for their performance. \textbf{RAG distance metrics:}~\label{sec:RAG-metrics}
%\newline
For given two vector embeddings, $\hat{u}_1$ and $\hat{u}_2$, and angle $\theta$ between the two, different distance metrics for RAG computation are given as:

\begin{equation}
    \text{cosine similarity} = 1 - cos(\theta)
\end{equation}

\begin{equation}
    \mathcal{L}_2 = \lvert \lvert \hat{u}_1 - \hat{u}_2 \lvert \lvert  
\end{equation}

\begin{equation}
    \text{dot product} = \hat{u}_1 \cdot \hat{u}_2
\end{equation}

Thus, the cosine similarity metric disregards the magnitude of the vectors and accounts for only for their relative orientation.

\subsection{Instruction structuring}
As illustrated in Fig.~\ref{fig:framework_pipeline}, a reinforced instructional structuring is adopted in this work. First, a standardized \verb|JSON| template is created for each task. Such a JSON template includes four primary attributes: (a) types of failures for a particular operation, (b) deterministic status of for that particular failure type, (c) a sub-failure for each type of failure object, and (d) the reason for each type of sub-failure. For instance, for an operation like palletization, types of failure objects could include instances such as robot-arm, loading pallet, flipping-station, etc. For this work, the deterministic status for failure objects have been constrained to `normal', `anomalous' or `unknown'. Within the scope of this work, it has been observed that such a constraint while limiting the response options for the VLM, significantly improves the performance (absolute accuracy) and also provides a concise metric for quantifying the model's performance. For each failure type, a sub-failures are specified attributes that can allow to identify the failure cause. For example, the failure of robot-arm could be attribute to failure of suction-gripper, or limitations of joint-motions due to motion singularities. Finally, the reasons for sub-failures are unconstrained natural language responses that can allow operators/inspectors to debug the failure scenario.

\section{Experiments}\label{sec:experiments}

\begin{figure}
    \centering
    \includegraphics[width=\columnwidth]{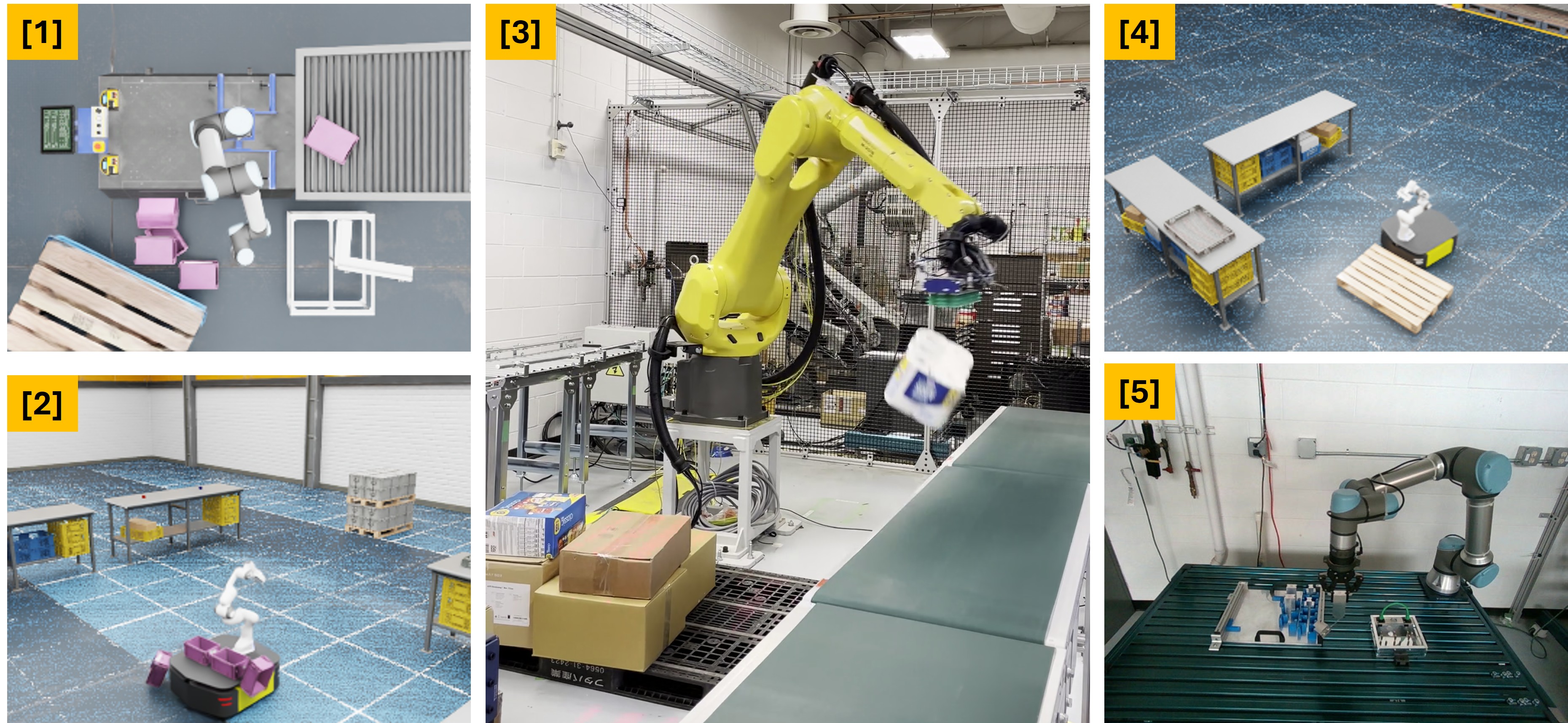}
    \caption{Five experiment setups in simulation ([1] Palletization, [2] MoMa - Transport, [4] MoMa - Traverse) and physical experiments ([3] Depalletization, [5] Assembly) used to evaluate proposed failure detection methods}
    \label{fig:experiments}
\end{figure}

We conducted simulation using two robot platforms for a set of tasks in Nvidia Isaac Sim: a fixed robot arm and a mobile manipulator. The fixed robot arm (UR5 from Universal Robots) is programmed for palletizing. The mobile manipulator (MoMa) consists of a mobile robot base (Clearpath Ridgeback) and a 7 DoF robot arm (Franka). Further more, we used a physical UR5 and a Fanuc M-20iB robot arm for assembly, cleaning and depalletizing.

\subsection{Robot operations}
Figure \ref{fig:experiments} shows the five experiment setups used to evaluate the proposed RAG-based failure detection methods. 

\textbf{1. Palletizing: } a UR5 robot is programmed to pick incoming boxes from the conveyor and place them on a pallet. Boxes are three-walled (one side open), the robot arm should place all boxes with the open side on the pallet. When an incoming box has its open side facing upwards, the robot should flip the box by using the L-shaped station next to the conveyor. We simulate this palletizing process and potential failures caused by: irregular box size, inaccurate L-shape station location.

%\textbf{MoMa - Pick-and-Place: } a MoMa is used for an order fulfillment task where inventory items are on the right bin and packing bin is on the left. The robot should pick items based on customer's order.

\textbf{2. MoMa - Transport: } a MoMa is transporting some goods inside a warehouse. The goods are placed on the mobile robot platform. Stack of goods might be unstable during transportation due to MoMa accelerates/decelerates. 

\textbf{3. Depalletizing: } a Fanuc robot (M-20iB) is depalletizing packages from a pallet to a moving conveyor. Packages include cardboard boxes and shrink-wrapped items. Soft plastic packaging material might cause picking failure or loose grasping.

\textbf{4. MoMa - Traverse: } a MoMa is moving from point A to point B after receiving the command from fleet management software. Some minor layout changes to the environment that haven't been updated to the map might cause collisions.

\textbf{5. Assembly: } a UR5 robot is used to assemble a mechanical kit by picking components from one tray and placing them into target locations. Inaccuracy in workstation setup might cause assembly errors.

%\textbf{Cleaning: } a UR5 robot cleans the table by placing items into target containers one by one.

\subsection{RAG Dataset}
% information provided --> images, json template (ground truth)
% images --> different frequencies

\begin{table}
\centering
\caption{Framework Attributes}
\label{tab:RAGDataset}
\renewcommand{\arraystretch}{1.3}
\begin{tabular}{lcc}
\hline
\textbf{Operation} & \textbf{Variations} & \textbf{Sampling Rate (Hz)} \\
\hline
Palletization        & $6$ & $0.05$ to $0.27$ \\
MoMa -- Transport    & $3$ & $0.5$ to $2.5$ \\
Depalletization      & $3$ & $0.5$ to $2.5$ \\
MoMa -- Traverse     & $2$ & $0.5$ to $2.5$ \\
Assembly             & $2$ & $0.2$ to $1$ \\
\hline
\multicolumn{3}{c}{\textbf{Inference Tasks}} \\
\hline
Vector Encoding & \multicolumn{2}{c}{CLIP : \texttt{ViT-B-32}} \\
VLM Inference   & \multicolumn{2}{c}{Qwen : \texttt{qwen2.5vl:32b}} \\
\hline
\end{tabular}
\end{table}

Table~\ref{tab:RAGDataset} gives an overview of the RAG dataset composed for failure identification. Every operation can have several failure scenarios apart from the baseline (expected) function. Further, each Operation-failure pair has been captured at different sampling frequencies. The idea behind capturing different frequencies is to provide insights on the tradeoff between the framework's real-time operational limits (too high sampling frequency may result in computational overload) vs information contained withing the query image (too low frequency may not capture relevant operation feature and may result in poor absolute performance/ absence of adequate reasoning). In this work, the frame sampling frequency ranges between $0.05$ Hz to $2.5$ Hz for different operations in net $5$ gradations. For every set \{\textbf{operation}, \textbf{failure scenario}, \textbf{sampling frequency} \}, $5$ instances of distinct images have been provided for improve the generalization of the framework. Thus, for every operation a total of $\textbf{failure scenarios} \times 5~(\textbf{sampling frequencies}) \times 5~(\textbf{instances})$ data points have been captured. The performance validation occurs on $3$ images not used for composing the RAG dataset, thus making a roughly $60-40$ training validation split.

While categorizing different processes via sampling frequencies makes sense from an engineering perspective, it is difficult to translate that from the LLM-processing point-of-view. From the language model perspective (including RAG) the difference in different failure scenarios and their corresponding visual representations can only be distinguished by feature similarities or differences, especially in the latent embedding sense. Thus, for improved interpretation of the VLM's performance, different images embeddings (for query images) are utilized as parametric variation and utilized to investigate the variability in the retrieval performance and the VLM's performance.

\subsection{VLM Models}~\label{sec:vlm-models}
Over the last couple of years, several competitive VLMs have emerged (both proprietary and open-source) that could potentially be utilized for this framework. For this work, the open-source \textbf{Qwen} family of models have be chosen for benchmarking. The choice of these models comes from : (a) The primary emphasis of the work being on improvements brought about by RAG augmentaion to off-the-shelf VLM models and not the standalone performance of different models, (b) the models being available in different parameter sizes (3b, 11b \& 32b) for a variable compute constraints, and, (c) being open-source and widely adaptable to different application programming interfaces such as `Ollama' to enable recreation of our benchmarks by the community. Table~\ref{tab:RAGDataset} illustrates the details of the different parameters and models utilized for the proposed framework.

\subsection{Results}~\label{sec:results}
\begin{figure}[t]
    \centering
    \includegraphics[width=\columnwidth]{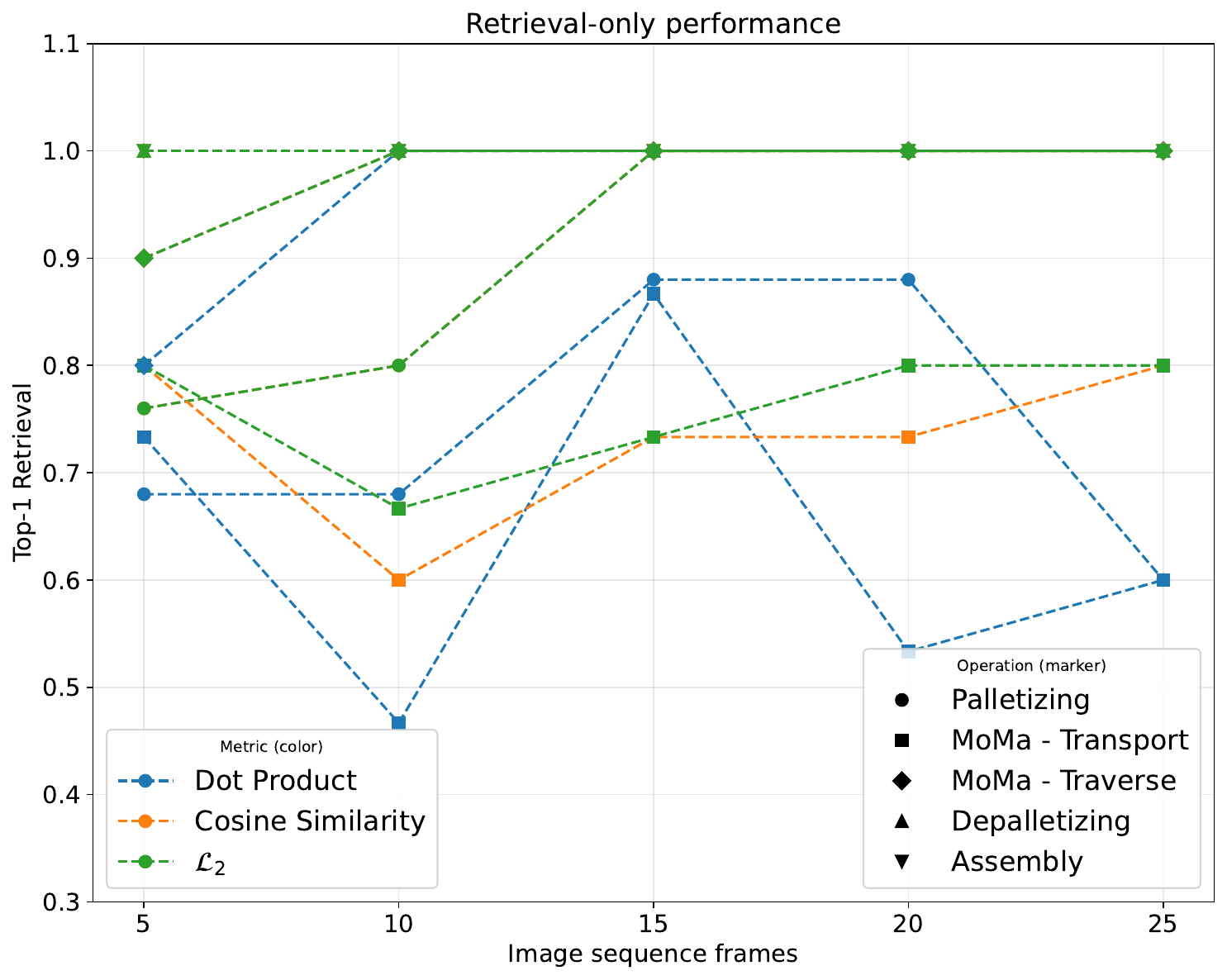}
    \caption{Retrieval-only performance captures the ability of different RAG measurement metrics~(\ref{sec:formulation}) such as (a) cosine similarity, (b)$\mathcal{L}_2$~distance, and (c) vector dot product in retrieving relevant failure cases.}
    \label{fig:retrieval-only-results}
\end{figure}

\begin{figure}[t]
    \centering
    \includegraphics[width=\columnwidth]{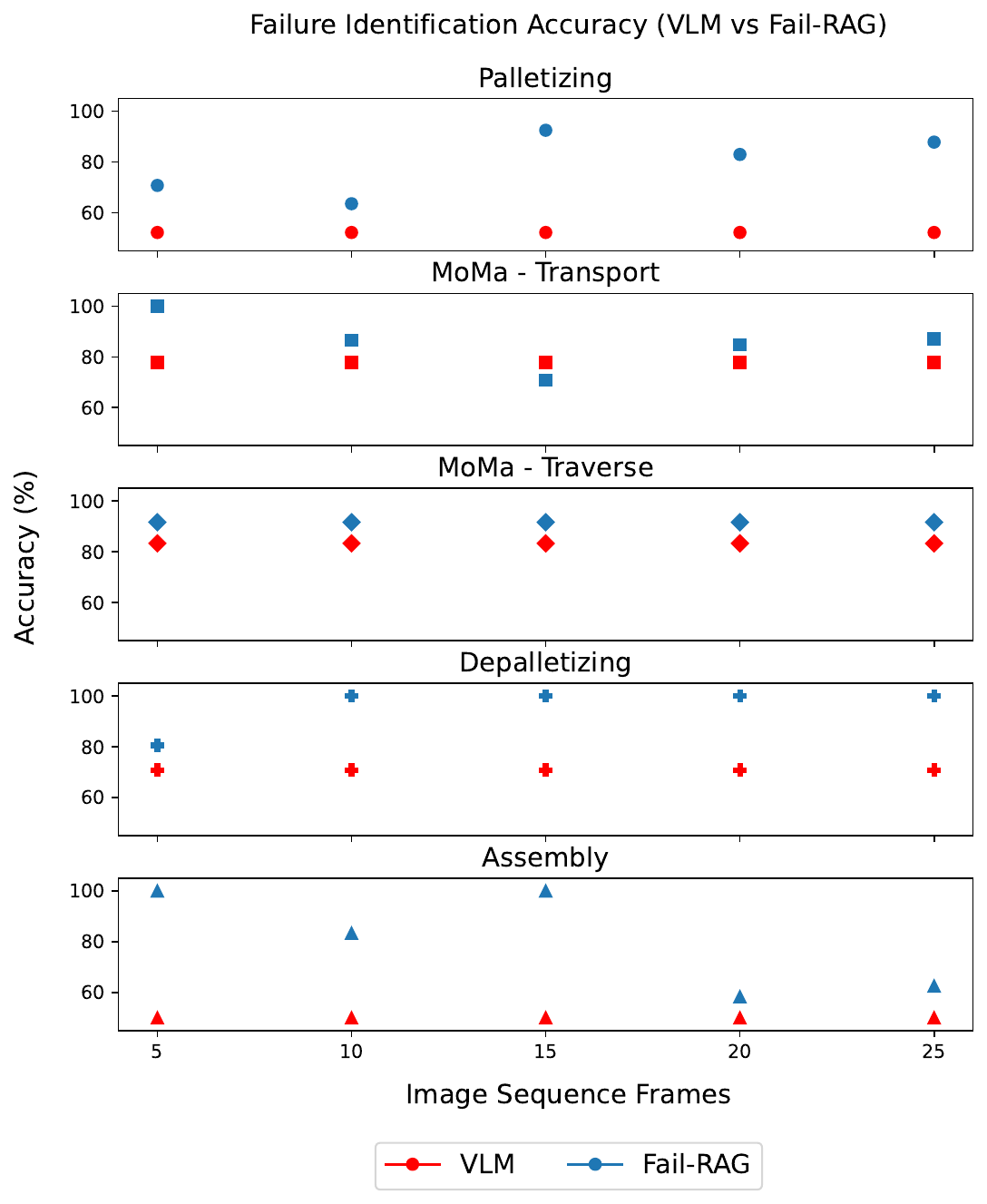}
    \caption{Comparison of the VLM and the proposed Fail-RAG framework for accurate failure identification. The metrics capture the correct characterization of several operation indicators such as \texttt{robot\_arm},~\texttt{payload},~\texttt{overall\_operation}, etc. as $\mathbf{normal}$ or $\mathbf{anomalous}$ from the agent's response. For all operations, Fail-RAG consistently shows better performance as compared to vanilla VLM, except for an outlier for one of the cases of MoMa-Transport.} 
    \label{fig:vlm-rag-results}
\end{figure}

Based on the experiments outlined in sec~\ref{sec:experiments}, the performance of the framework has been outline in two steps. The RAG-only performance captures how different RAG metrics work in pure retrieval of the relevant RAG data given the query image. This is purely mathematical, and hallucination free performance evaluation to investigate image retrieval. In the second step, VLM-RAG is compared against pure VLM.

\subsubsection{Retrieval Performance}

As outlined in section~\ref{sec:RAG-metrics}, the retrieval only performance is evaluated across all operations and their failure scenarios for all three RAG distance metrics. Figure~\ref{fig:retrieval-only-results} illustrates this performance with the sampling frequency for different task increasing across the horizontal axis. The figure also utilize distinct colors and markers for differentiating between different distance measurement metric and different robot operations respectively. The retrieval scores ranging from $0$ to $1$ indicate the mean score for identifying different queries which individually return $0$ and $1$. For instance, if $6$ out of $10$ queried images return the correct failure mode, the score is $0.6$. 

The evaluations indicated that more than a specific distance measurement metric, or a specific frame rate for data collection across all operation, it is a combination of a distance metirc and operation that may create variable performance. In general, for all operations, $\mathcal{L}_2$ distance and cosine similarity work quiet well irrespective of the sampling frequencies. Within different robot operations, while most other operations show expected performance, \textbf{MoMa - Transport} shows poor performance across all distance measurement metric as well as different sampling frequencies. Based on the results cosine distance is chosen as the preferred method to measure the distance based for RAG framework as it allows for equivalent performance for vector embeddings based on images and texts alike.

\subsubsection{VLM-RAG}

Figure~\ref{fig:vlm-rag-results} illustrates the comparative performance between the the VLM model (baseline) and Fail-RAG for all different operations and across different frame rates. The evaluations capture the accuracy of the agent in identifying not only if the image indicates `failure', but also the type of failure as provided in the in the RAG documents. For example, for an operation such as Palletizing, different sub-failure modes may include \verb|robot_arm|, \verb|pallet|, \verb|flipping_station|, and, \verb|overall_operation| along with the corresponding reasons. The evaluations illustrated in Fig.~\ref{fig:vlm-rag-results} capture if the agent identifies correctly if the sub-failures are \textbf{normal} or \textbf{anomalous}.

Results indicate in general an average of around $25\%$ increase in the accuracy performance across all operations with a few increasing by as much as $40\%$. Fail-RAG outperforms significantly for all operations expect for a single scenario for \textbf{MoMa - Transport}. Compared to the simulation scenarios, results are much more improved over the baseline for real experiments such as \textbf{Depalletizing} and \textbf{Assembly}. One potential reason for this is that as compared to simulation, there is more pronounced impact of shadows and lighting in physical experiments, thus making RAG more relevant.

Unfortunately, there wasn't any noticeable pattern for Fail-RAG's performance for all operations as function of increasing/decreasing the frame rate. While there were variable results for individual operations (such as the best performance for \textbf{Palletizing} was for $15$ frames and for \textbf{MoMa-Transport} was for $5$ frames), there was no common trend across all operations. This lead to analysis from the the perspective of vector embeddings used within the RAG database.

\section{Analysis}~\label{sec:analysis}

\begin{figure*}
    \centering
    \includegraphics[width=\textwidth]{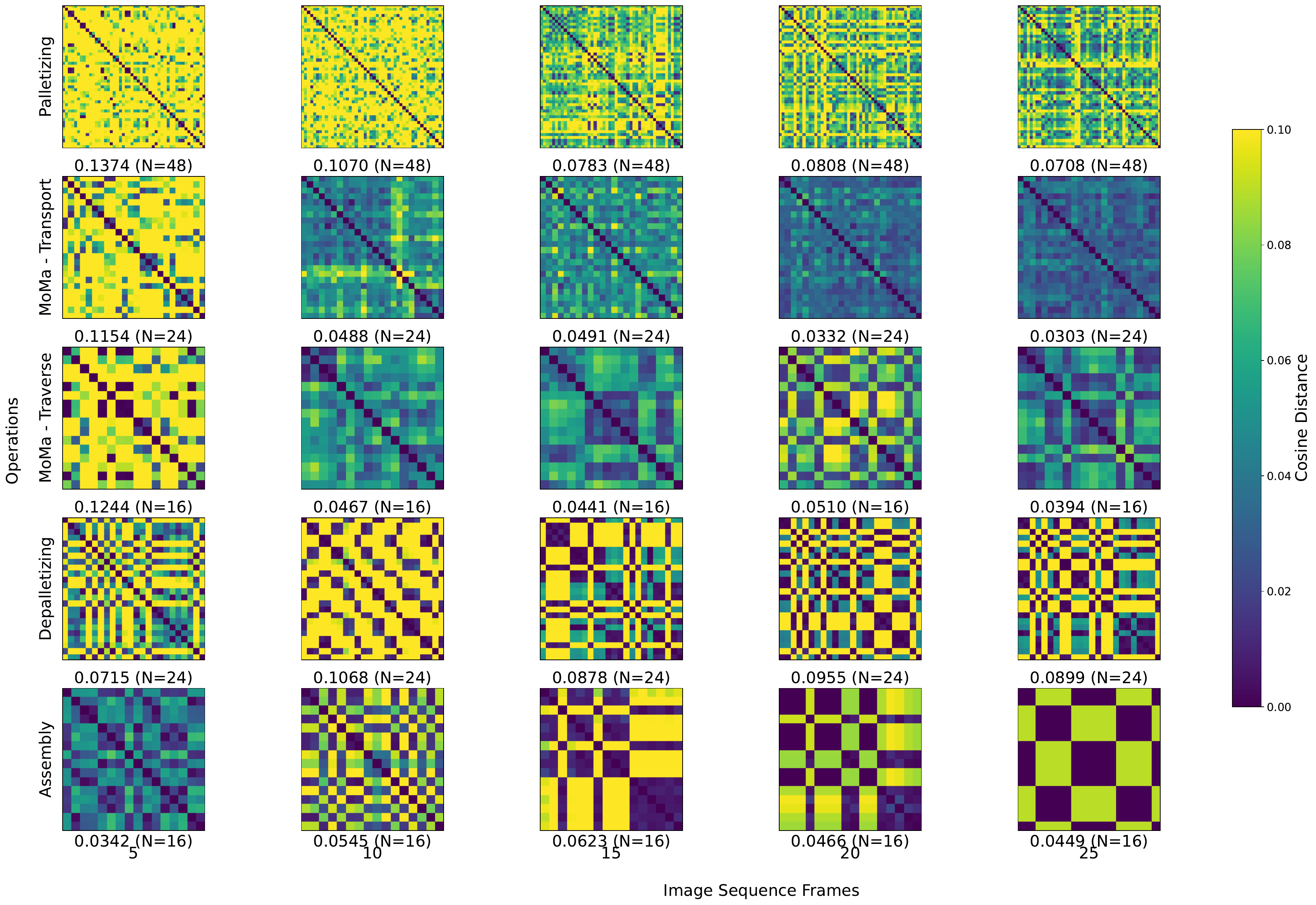}
    \caption{Heatmaps illustrating the cosine distance between any two vector embeddings of an operation. For every operation, there are $8$ distinct vectors for every scenario. Thus, total number of vectors per tile is $N = 8 \times \text{scenarios}$ . For a given set of $N$ vectors, the horizontal and the vertical axes aligns them inversely such that the diagonal is the the same vector comparing with itself resulting all diagonal values to be $0.00$. Tiles with dense yellow hue correspond to operation - image sequence combinations where the scenario embeddings are very distinct from one another where are the ones with blue hue are the ones with vector embeddings similar to each other. Similarity is measured by cosine distance ($1 - \text{cosine similarity}$) which indicates $0.0$ for identical vectors and $1.0$ when vectors are orthogonal. Below each tile is mean cosine distance between all the vectors for that operation-image sequence combination along with $N$, the total number of vectors used for averaging.}
    \label{fig:mean_cos_dist}
\end{figure*}

\begin{figure}
    \centering
    \includegraphics[width=\columnwidth]{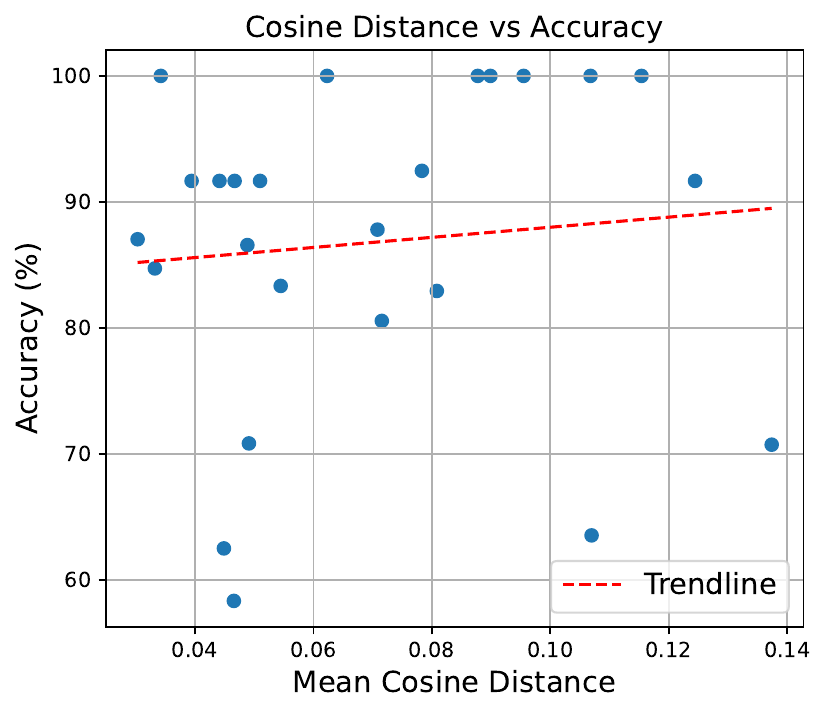}
    \caption{The accuracy retrieved from the Fail-RAG evaluations (Fig.~\ref{fig:vlm-rag-results}) can be arranged against the mean cosine distance calculations illustrated in Fig.~\ref{fig:mean_cos_dist}. Such a perspective now maps how the Fail-RAG performance varies as the mean cosine distance changes. As expected, increasing the cosine distance (which indicates failure scenarios are encoded distinctly from one another) results in improved Fail-RAG performance.}
    \label{fig:dist_acc_trend}
\end{figure}

Within the RAG framework, every image and text is encoded and stored as a $\mathbb{R}^{512}$ vector. When presented with a query image the CLIP model encodes the query image as $\mathbb{R}^{512}$ and uses the distance metric (cosine similarity) to retrieve the vectors mostly aligned with the query vector. For a RAG database containing several failure scenarios of a robot operation as a vector embeddings, we hypothesize that the RAG performance can be dictated based on how distinct the vector embeddings for each failure scenario can be distinct from one another. 

Figure~\ref{fig:mean_cos_dist} illustrates the phenomenon of comparing vectors within a database with one another. The $25$ tiles indicate $25$ distinct RAG databases for the image embeddings. Each tile represents a set of $N$ vectors measured against the same $N$ vectors arranged inversely. This results in all diagonal measurements being the measurements between identical vectors rendering the distance to be $0.0$. The hues yellow and blue indicate how distinct and similar the vectors are from each other respectively. Thus, a tile with dense yellow hue indicates that all vectors are very distinct from each other and one with deep blue hues indicate that all vectors are quite similar to each other. Furthermore, the pixel density is based on the total number of embeddings in the database. Thus, tiles with higher number of distinct vectors ($N = 48$) have higher pixel density relative to the ones with lower vector count ($N = 16$).

When treating every operation independently, and comparing with the results of Fig.~\ref{fig:vlm-rag-results}, it can be observed that in general higher mean cosine distance is indicative of higher Fail-RAG performance. While there are some outliers to the trend, the general trend indicates alignment with our proposed hypothesis. Figure~\ref{fig:dist_acc_trend} illustrates this effect more explicitly. The mean cosine distances (one for each tile) and corresponding Fail-RAG performance are plotted against one another. The trendline now indicates a general trend in which increasing cosine distance improves the RAG based VLM performance. 

\section{Discussion}\label{sec:discussion}
In this work, Fail-RAG : a RAG based VLM framework has been proposed towards failure identification in robotics processes. Fail-RAG is significantly effort-efficient as compared to the majority of the contemporary works which focus on fine-tuning VLMs/VLAs with failure-datasets. Furthermore, this work presents a framework for curating failure data samples and provides insights on how different data collection procedures may result in performance variability. Fail-RAG also demonstrates significant performance improvement over the baseline demonstrating the utility of off-the-shelf VLMs without any expensive fine-tuning.

More critically, we emphasize on diving deeper on internal mechanics of the RAG framework to investigate the sensitivity of the RAG framework as a function of how the vector embeddings are encoded in the dataset, how their similarity is measured, and how their distinct directions in the embedding space impacts the framework's performance. The interesting implications of this effort now scales beyond just robot failure identification and can potentially be adopted for general purpose vision based monitoring and inspection beyond robotics.

While composing several frames together and presenting a single query image makes the framework lightweight, it also introduces some delay in identification of the failure. Depending on the task, such a delay might be acceptable/ undesirable due to potential disruptions to automation workflows. Thus, minimizing the delay can be a potential future direction to build on top of this work. Working in the latent space or/and introducing new operational views can be some methods that could be adopted towards early prediction of the failure thus reducing the identification delay.

Furthermore, while figure~\ref{fig:mean_cos_dist} illustrates that variability in vector embeddings is crucial for improving performance, the exact mathematical modeling or formalizing the relation between them can further strengthen the adoption of this framework. Such a relationship can now be utilize to curate the RAG dataset thus improving the efficiency of the data-collection process.


%%%%%%%%%%%%%%%%%%%%%%%%%%%%%%%%%%%%%%%%%%%%%%%%%%%%%%%%%%%%%%%%%%%%%%%%%%%%%%%%

%%%%%%%%%%%%%%%%%%%%%%%%%%%%%%%%%%%%%%%%%%%%%%%%%%%%%%%%%%%%%%%%%%%%%%%%%%%%%%%%

%%%%%%%%%%%%%%%%%%%%%%%%%%%%%%%%%%%%%%%%%%%%%%%%%%%%%%%%%%%%%%%%%%%%%%%%%%%%%%%%
%\section*{APPENDIX}
%\include{appendix_1}
% \section*{ACKNOWLEDGMENT}

%\printbibliography
\bibliographystyle{IEEEtran}
\bibliography{VLM-FailureDetection,ref}
%%%%%%%%%%%%%%%%%%%%%%%%%%%%%%%%%%%%%%%%%%%%%%%%%%%%%%%%%%%%%%%%%%%%%%%%%%%%%%%%
\end{document}